\documentclass{article}

\usepackage{microtype}
\usepackage{graphicx}
\usepackage{subfigure}
\usepackage{booktabs} 
\usepackage[sort&compress]{natbib}
\usepackage{amssymb}
\usepackage{mathtools}
\usepackage{bm}
\usepackage{cancel}
\usepackage{listings}

\usepackage{hyperref}

\newcommand{\AlgoName}{\textsc{Barlow Twins}}
\newcommand{\norm}[1]{\left\lVert#1\right\rVert}

\usepackage{xcolor}

\usepackage{xspace}
\usepackage{enumitem}
\usepackage{multirow}
\usepackage{array}
\usepackage{cleveref}
\makeatletter
\DeclareRobustCommand\onedot{\futurelet\@let@token\@onedot}
\def\@onedot{\ifx\@let@token.\else.\null\fi\xspace}
\def\eg{\emph{e.g}\onedot} 
\def\ie{\emph{i.e}\onedot}

\crefname{section}{\S}{\S\S}
\crefname{subsection}{\S}{\S\S}
\crefformat{table}{Table~#2#1#3}
\crefformat{figure}{Fig~#2#1#3}
\crefformat{equation}{Eq~#2#1#3}

\usepackage[accepted]{icml2021}

\icmltitlerunning{Barlow Twins: Self-Supervised Learning via Redundancy Reduction}

\begin{document}

\twocolumn[
\icmltitle{Barlow Twins: Self-Supervised Learning via Redundancy Reduction}



\icmlsetsymbol{equal}{*}

\begin{icmlauthorlist}
\icmlauthor{Jure Zbontar}{equal,fb}
\icmlauthor{Li Jing}{equal,fb}
\icmlauthor{Ishan Misra}{fb}
\icmlauthor{Yann LeCun}{fb,nyu}
\icmlauthor{Stéphane Deny}{fb}
\end{icmlauthorlist}

\icmlaffiliation{fb}{Facebook AI Research}
\icmlaffiliation{nyu}{New York University, NY, USA}

\icmlcorrespondingauthor{Jure Zbontar}{jzb@fb.com}
\icmlcorrespondingauthor{Li Jing}{ljng@fb.com}
\icmlcorrespondingauthor{Ishan Misra}{imisra@fb.com}
\icmlcorrespondingauthor{Yann LeCun}{yann@fb.com}
\icmlcorrespondingauthor{Stéphane Deny}{stephane.deny.pro@gmail.com}

\icmlkeywords{Machine Learning, ICML}

\vskip 0.3in
]



\printAffiliationsAndNotice{\icmlEqualContribution} 

\begin{abstract}

Self-supervised learning (SSL) is rapidly closing the gap with supervised methods on large computer vision benchmarks. A successful approach to SSL is to learn embeddings which are invariant to distortions of the input sample. However, 
a recurring issue with this approach is the existence of trivial constant solutions. Most current methods avoid such solutions by careful implementation details. We propose an objective function that naturally avoids collapse by measuring the cross-correlation matrix between the outputs of two identical networks fed with distorted versions of a sample, and making it as close to the identity matrix as possible. This causes the embedding vectors of distorted versions of a sample to be similar, while minimizing the redundancy between the components of these vectors. The method is called \AlgoName{}, owing to neuroscientist H. Barlow's \emph{redundancy-reduction principle} applied to a pair of identical networks. \AlgoName{} does not require large batches nor asymmetry between the network twins such as a predictor network, gradient stopping, or a moving average on the weight updates. Intriguingly it benefits from very high-dimensional output vectors. \AlgoName{} outperforms previous methods on ImageNet for semi-supervised classification in the low-data regime, and is on par with current state of the art for ImageNet classification with a linear classifier head, and for transfer tasks of classification and object detection.\footnote{Code and pre-trained models (in PyTorch) are available at https://github.com/facebookresearch/barlowtwins}
\enlargethispage{\baselineskip}

\end{abstract}

\section{Introduction}

\begin{figure}[ht]
\vskip 0.2in
\begin{center}
\centerline{\includegraphics[width=9cm]{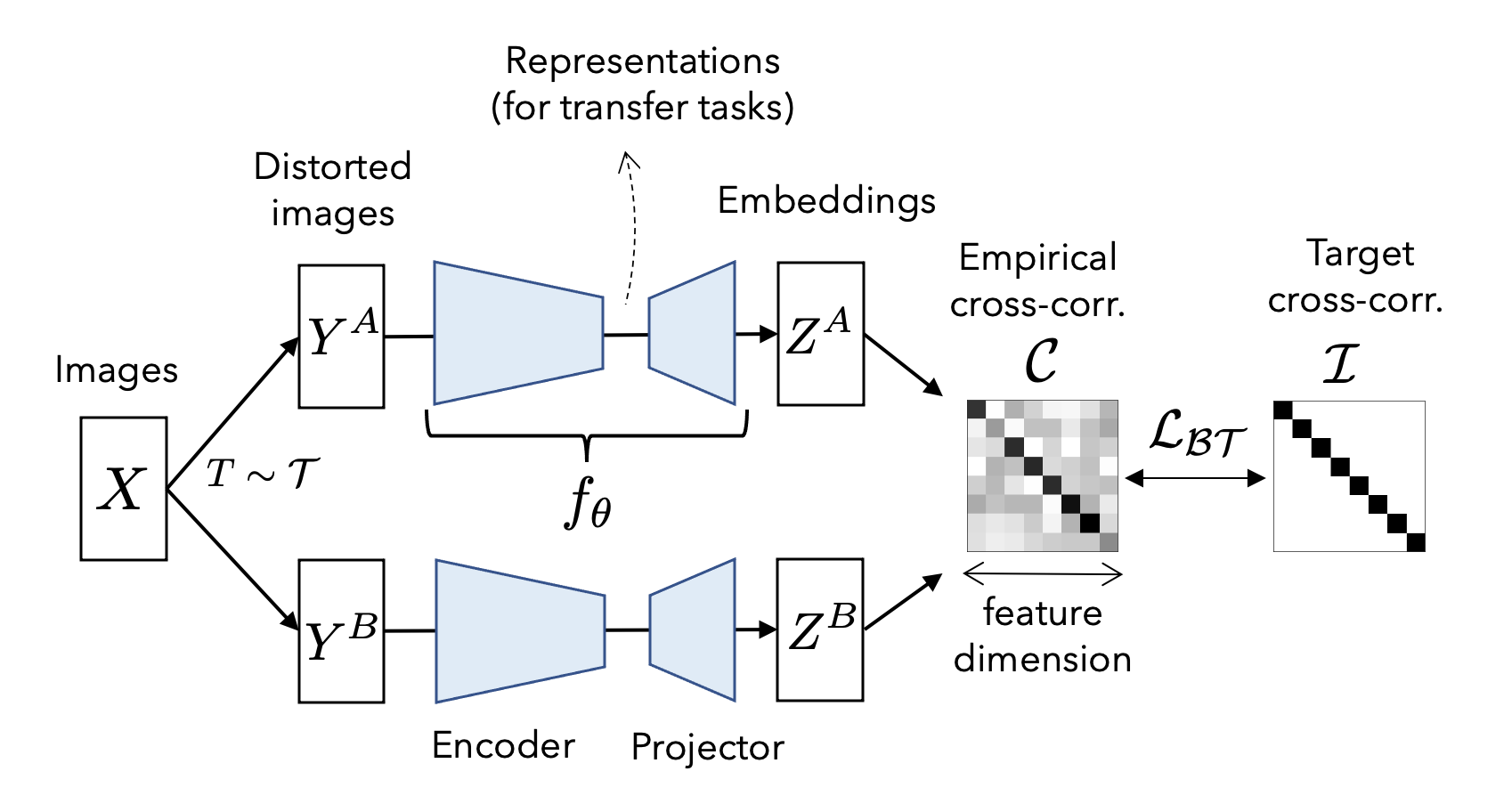}}
\caption{\AlgoName{}'s objective function measures the cross-correlation matrix between the embeddings of two identical networks fed with distorted versions of a batch of samples, and tries to make this matrix close to the identity. This causes the embedding vectors of distorted versions of a sample to be similar, while minimizing the redundancy between the components of these vectors. \AlgoName{} is competitive with state-of-the-art methods for self-supervised learning while being conceptually simpler, naturally avoiding trivial constant (i.e. collapsed) embeddings, and being robust to the training batch size.}
\label{fig:fig_method}
\end{center}
\vskip -0.2in
\end{figure}

Self-supervised learning aims to learn useful representations of the input data without relying on human annotations. 
Recent advances in self-supervised learning for visual data~\cite{caron2020swav,grill2020bootstrap,chen2020simple,he2019momentum,misra2019self} show that it is possible to learn self-supervised representations that are competitive with supervised representations. 
A common underlying theme that unites these methods is that they all aim to learn representations that are invariant under different distortions (also referred to as `data augmentations'). 
This is typically achieved by maximizing similarity of representations obtained from different distorted versions of a sample using a variant of Siamese networks~\cite{hadsell2006dimensionality}.
As there are trivial solutions to this problem, like a constant representation, these methods rely on different mechanisms to learn useful representations.

Contrastive methods like \textsc{SimCLR}~\cite{chen2020simple} define `positive' and `negative' sample pairs which are treated differently in the loss function.
Additionally, they can also use asymmetric learning updates wherein momentum encoders~\cite{he2019momentum} are updated separately from the main network. 
Clustering methods use one distorted sample to compute `targets' for the loss, and another distorted version of the sample to predict these targets, followed by an alternate optimization scheme like k-means in \textsc{DeepCluster}~\cite{caron2018deep} or non-differentiable operators in \textsc{SwAV}~\cite{caron2020swav} and \textsc{SeLa}~\cite{asano2019self}.
In another recent line of work, \textsc{BYOL}~\cite{grill2020bootstrap} and \textsc{SimSiam}~\cite{chen2020exploring}, both the network architecture and parameter updates are modified to introduce asymmetry. The network architecture is modified to be asymmetric using a special `predictor' network and the parameter updates are asymmetric such that the model parameters are only updated using one distorted version of the input, while the representations from another distorted version are used as a fixed target.
~\cite{chen2020exploring} conclude that the asymmetry of the learning update, `stop-gradient', is critical to preventing trivial solutions.

In this paper, we propose a new method, \AlgoName{}, which applies \emph{redundancy-reduction} --- a principle first proposed in neuroscience --- to self-supervised learning. In his influential article \emph{Possible Principles Underlying the Transformation of Sensory Messages} \citep{barlow_possible_nodate}, neuroscientist H. Barlow hypothesized that the goal of sensory processing is to recode highly redundant sensory inputs into a factorial code (a code with statistically independent components). This principle has been fruitful in explaining the organization of the visual system, from the retina to cortical areas (see \citep{barlow_redundancy_2001} for a review and \citep{ocko_emergence_2018,lindsey_unified_2020,schwartz_natural_2001} for recent efforts), and has led to a number of algorithms for supervised and unsupervised learning \citep{redlich_redundancy_1993,redlich_supervised_1993,deco_non-linear_1997,foldiak_forming_1990,linsker_self-organization_1988,schmidhuber_semilinear_1996,balle_end--end_2017}. Based on this principle, we propose an objective function which tries to make the cross-correlation matrix computed from twin embeddings as close to the identity matrix as possible. \AlgoName{} is conceptually simple, easy to implement and learns useful representations as opposed to trivial solutions. Compared to other methods, it does not require large batches \cite{chen2020simple}, nor does it require any asymmetric mechanisms like prediction networks \cite{grill2020bootstrap}, momentum encoders \cite{he2019momentum}, non-differentiable operators \cite{caron2020swav} or stop-gradients \cite{chen2020exploring}. Intriguingly, \AlgoName{} strongly benefits from the use of very high-dimensional embeddings. \AlgoName{} outperforms previous methods on ImageNet for semi-supervised classification in the low-data regime (55\% top-1 accuracy for 1\% labels), and is on par with current state of the art for ImageNet classification with a linear classifier head, as well as for a number of transfer tasks of classification and object detection. 

\section{Method}

\subsection{Description of \AlgoName{}}



Like other methods for SSL \cite{caron2020swav,grill2020bootstrap,chen2020simple,he2019momentum,misra2019self}, \AlgoName{} operates on a joint embedding of distorted images (Fig. \ref{fig:fig_method}). More specifically, it produces two distorted views for all images of a batch $X$ sampled from a dataset. The distorted views are obtained via a distribution of data augmentations $\mathcal{T}$. The two batches of distorted views $Y^A$ and $Y^B$ are then fed to a function $f_{\theta}$, typically a deep network with trainable parameters $\theta$, producing batches of embeddings $Z^{A}$ and $Z^{B}$ respectively. To simplify notations, $Z^{A}$ and $Z^{B}$ are assumed to be mean-centered along the batch dimension, such that each unit has mean output 0 over the batch. 

\AlgoName{} distinguishes itself from other methods by its innovative loss function $\mathcal{L_{BT}}$:

\begin{equation}
\mathcal{L_{BT}} \triangleq  \underbrace{\sum_i  (1-\mathcal{C}_{ii})^2}_\text{invariance term}  + ~~\lambda \underbrace{\sum_{i}\sum_{j \neq i} {\mathcal{C}_{ij}}^2}_\text{redundancy reduction term}
\label{eq:lossBarlow}
\end{equation}

where $\lambda$ is a positive constant trading off the importance of the first and second terms of the loss, and where $\mathcal{C}$ is the cross-correlation matrix computed between the outputs of the two identical networks along the batch dimension:

\begin{equation}
\mathcal{C}_{ij} \triangleq \frac{
\sum_b z^A_{b,i} z^B_{b,j}}
{\sqrt{\sum_b {(z^A_{b,i})}^2} \sqrt{\sum_b {(z^B_{b,j})}^2}}
\label{eq:crosscorr}
\end{equation}

where $b$ indexes batch samples and $i,j$ index the vector dimension of the networks' outputs. $\mathcal{C}$ is a square matrix with size the dimensionality of the network's output, and with values comprised between -1 (i.e. perfect anti-correlation) and 1 (i.e. perfect correlation). 


Intuitively, the \emph{invariance term} of the objective, by trying to equate the diagonal elements of the cross-correlation matrix to 1, makes the embedding invariant to the distortions applied.  The \emph{redundancy reduction term}, by trying to equate the off-diagonal elements of the cross-correlation matrix to 0, decorrelates the different vector components of the embedding. This decorrelation reduces the redundancy between output units, so that the output units contain non-redundant information about the sample. 

More formally, \AlgoName{}'s objective function can be understood through the lens of information theory, and specifically as an instanciation of the \emph{Information Bottleneck (IB)} objective \cite{tishby_deep_2015,tishby_information_2000}. Applied to self-supervised learning, the IB objective consists in finding a representation that conserves as much information about the sample as possible while being the \emph{least} possible informative about the specific distortions applied to that sample. The mathematical connection between \AlgoName{}'s objective function and the IB principle is explored in Appendix A.

\AlgoName{}' objective function has similarities with existing objective functions for SSL. For example, the redundancy reduction term plays a role similar to the \emph{contrastive term} in the \textsc{infoNCE} objective \cite{oord2018representation}, as discussed in detail in Section 5. However, important conceptual differences in these objective functions result in practical advantages of our method compared to \textsc{infoNCE}-based methods, namely that (1) our method does not require a large number of negative samples and can thus operate on small batches (2) our method benefits from very high-dimensional embeddings. Alternatively, the redundancy reduction term can be viewed as a \emph{soft-whitening} constraint on the embeddings, connecting our method to a recently proposed method performing a \emph{hard-whitening} operation on the embeddings \cite{ermolov_whitening_2020}, as discussed in Section 5. However, our method performs better than current hard-whitening methods.

The pseudocode for \AlgoName{} is shown as Algorithm~\ref{alg:barlow_twins}.

\subsection{Implementation Details}
\label{sec:implementation_details}

\begin{algorithm}[tb]
   \caption{PyTorch-style pseudocode for Barlow Twins.}
   \label{alg:barlow_twins}
   
    \definecolor{codeblue}{rgb}{0.25,0.5,0.5}
    \lstset{
      basicstyle=\fontsize{7.2pt}{7.2pt}\ttfamily\bfseries,
      commentstyle=\fontsize{7.2pt}{7.2pt}\color{codeblue},
      keywordstyle=\fontsize{7.2pt}{7.2pt},
    }
\begin{lstlisting}[language=python]
# f: encoder network
# lambda: weight on the off-diagonal terms
# N: batch size
# D: dimensionality of the embeddings
#
# mm: matrix-matrix multiplication
# off_diagonal: off-diagonal elements of a matrix
# eye: identity matrix

for x in loader: # load a batch with N samples
    # two randomly augmented versions of x
    y_a, y_b = augment(x)
    
    # compute embeddings
    z_a = f(y_a) # NxD
    z_b = f(y_b) # NxD
    
    # normalize repr. along the batch dimension
    z_a_norm = (z_a - z_a.mean(0)) / z_a.std(0) # NxD
    z_b_norm = (z_b - z_b.mean(0)) / z_b.std(0) # NxD
    
    # cross-correlation matrix
    c = mm(z_a_norm.T, z_b_norm) / N # DxD
    
    # loss
    c_diff = (c - eye(D)).pow(2) # DxD
    # multiply off-diagonal elems of c_diff by lambda
    off_diagonal(c_diff).mul_(lambda)
    loss = c_diff.sum()

    # optimization step
    loss.backward()
    optimizer.step()
\end{lstlisting}
\end{algorithm}

\paragraph{Image augmentations} Each input image is transformed twice to produce the two distorted views shown in Figure~\ref{fig:fig_method}. The image augmentation pipeline consists of the following transformations: random cropping, resizing to $224 \times 224$, horizontal flipping, color jittering, converting to grayscale, Gaussian blurring, and solarization. The first two transformations (cropping and resizing) are always applied, while the last five are applied randomly, with some probability. This probability is different for the two distorted views in the last two transformations (blurring and solarization). We use the same augmentation parameters as \textsc{BYOL}~\cite{grill2020bootstrap}.

\paragraph{Architecture} 
The encoder consists of a ResNet-50 network~\cite{he2016deep} (without the final classification layer, 2048 output units) followed by a projector network. The projector network has three linear layers, each with 8192 output units. The first two layers of the projector are followed by a batch normalization layer and rectified linear units. We call the output of the encoder the 'representations' and the output of the projector the 'embeddings'. The representations are used for downstream tasks and the embeddings are fed to the loss function of \AlgoName{}.

\paragraph{Optimization} 
We follow the optimization protocol described in \textsc{BYOL}~\cite{grill2020bootstrap}. We use the LARS optimizer~\cite{you2017large} and train for 1000 epochs with a batch size of 2048. We however emphasize that our model works well with batches as small as 256 (see Ablations). We use a learning rate of 0.2 for the weights and 0.0048 for the biases and batch normalization parameters. We multiply the learning rate by the batch size and divide it by 256. We use a learning rate warm-up period of 10 epochs, after which we reduce the learning rate by a factor of 1000 using a cosine decay schedule~\cite{loshchilov2016sgdr}. We ran a search for the trade-off parameter $\lambda$ of the loss function and found the best results for $\lambda = 5\cdot10^{-3}$. We use a weight decay parameter of $1.5 \cdot 10^{-6}$. The biases and batch normalization parameters are excluded from LARS adaptation and weight decay. Training is distributed across 32 V100 GPUs and takes approximately 124 hours. For comparison, our reimplementation of \textsc{BYOL} trained with a batch size of 4096 takes 113 hours on the same hardware.
\section{Results}

We follow standard practice~\cite{goyal2019scaling} and evaluate our representations by transfer learning to different datasets and tasks in computer vision.
Our network is pretrained using self-supervised learning on the training set of the ImageNet ILSVRC-2012 dataset \cite{deng2009imagenet} (without labels). We evaluate our model on a variety of tasks such as image classification and object detection, and using fixed representations from the network or finetuning it. We provide the hyperparameters for all the transfer learning experiments in the Appendix.

\subsection{Linear and Semi-Supervised Evaluations on ImageNet}

\paragraph{Linear evaluation on ImageNet}

We train a linear classifier on ImageNet on top of fixed representations of a ResNet-50 pretrained with our method. The top-1 and top-5 accuracies obtained on the ImageNet validation set are reported in Table~\ref{tab:abl-imagenet}. Our method obtains a top-1 accuracy of $73.2\%$ which is comparable to the state-of-the-art methods.

\begin{table}[ht]
\caption{\textbf{Top-1 and top-5 accuracies (in \%) under linear evaluation on ImageNet}. All models use a ResNet-50 encoder. Top-3 best self-supervised methods are \underline{underlined}.} 
\label{tab:abl-imagenet}
\vskip 0.15in
\begin{center}
\begin{tabular}{lcc}
\toprule
Method  &  Top-1 & Top-5 \\
\midrule
Supervised & 76.5 & \\
\midrule
\textsc{MoCo}    & 60.6 & \\
\textsc{PIRL} & 63.6 & - \\
\textsc{SimCLR} & 69.3  & 89.0 \\
\textsc{MoCo v2} & 71.1 & 90.1 \\
\textsc{SimSiam} & 71.3 & - \\
\textsc{SwAV} (w/o multi-crop) & 71.8 & - \\
\textsc{BYOL} & \underline{74.3} & 91.6  \\
\textsc{SwAV} & \underline{75.3} & - \\
\AlgoName{} (ours) & \underline{73.2} & 91.0 \\
\bottomrule
\end{tabular}
\end{center}
\vskip -0.1in
\end{table}

\paragraph{Semi-supervised training on ImageNet}
We fine-tune a ResNet-50 pretrained with our method on a subset of ImageNet. We use subsets of size $1\%$ and $10\%$ using the same split as \textsc{SimCLR}. The semi-supervised results obtained on the ImageNet validation set are reported in Table~\ref{tab:abl-semisupervised}. Our method is either on par (when using $10\%$ of the data) or slightly better (when using $1\%$ of the data) than competing methods.

\begin{table}[ht]
\caption{\textbf{Semi-supervised learning on ImageNet} using 1\% and 10\% training examples. Results for the supervised method are from~\cite{zhai2019s4l}. Best results are in \textbf{bold}.}
\label{tab:abl-semisupervised}
\vskip 0.15in
\begin{center}
\begin{tabular}{lcccc}
\toprule
Method     & \multicolumn{2}{c}{Top-$1$} & \multicolumn{2}{c}{Top-$5$} \\
\cmidrule(lr){2-3}\cmidrule(lr){4-5}
          &  1\% & 10\%            & 1\% & 10\% \\
\midrule
Supervised         & 25.4 & 56.4 & 48.4 & 80.4 \\
\midrule
\textsc{PIRL}               & - & -           & 57.2 & 83.8\\
\textsc{SimCLR}    & 48.3 & 65.6 & 75.5 & 87.8\\
\textsc{BYOL}      & 53.2 & 68.8 & 78.4 & 89.0\\
\textsc{SwAV}               & 53.9 & \bf{70.2} & 78.5 & \bf{89.9}\\
\AlgoName{} (ours) & \bf{55.0} & 69.7 & \bf{79.2} & 89.3\\
\bottomrule
\end{tabular}
\end{center}
\vskip -0.1in
\end{table}

\subsection{Transfer to other datasets and tasks}

\begin{table}[ht]
\caption{\textbf{Transfer learning: image classification.} We benchmark learned representations on the image classification task by training linear classifiers on fixed features. We report top-1 accuracy on Places-205 and iNat18 datasets, and classification mAP on VOC07. Top-3 best self-supervised methods are underlined.}
\label{tab:linear_transfer}
\vskip 0.15in
\begin{center}
\begin{tabular}{@{}lccc@{}}
\toprule
Method & Places-205 & VOC07 & iNat18 \\
\midrule
Supervised & 53.2 & 87.5 & 46.7\\
\midrule
SimCLR & 52.5 & 85.5 & 37.2 \\
MoCo-v2 & 51.8 & \underline{86.4} & 38.6 \\
SwAV (w/o multi-crop) & 52.8 & \underline{86.4} & 39.5 \\
SwAV & \underline{56.7} & \underline{88.9} & \underline{48.6} \\
BYOL & \underline{54.0} & \underline{86.6} & \underline{47.6} \\
\AlgoName{} (ours) & \underline{54.1} & 86.2 & \underline{46.5} \\
\bottomrule
\end{tabular}
\end{center}
\vskip -0.1in
\end{table}

\par \noindent \textbf{Image classification with fixed features} We follow the setup from~\cite{misra2019self} and train a linear classifier on fixed image representations, \ie, the parameters of the ConvNet remain unchanged. We use a diverse set of datasets for this evaluation - Places-205~\cite{zhou2014learning} for scene classification, VOC07~\cite{everingham2010pascal} for multi-label image classification, and iNaturalist2018~\cite{van2018inaturalist} for fine-grained image classification. We report our results in Table~\ref{tab:linear_transfer}. \AlgoName{} performs competitively against prior work, and outperforms SimCLR and MoCo-v2 on most datasets.

\par \noindent \textbf{Object Detection and Instance Segmentation} We evaluate our representations for the localization based tasks of object detection and instance segmentation. We use the VOC07+12~\cite{everingham2010pascal} and COCO~\cite{lin2014microsoft} datasets following the setup in~\cite{he2019momentum} which finetunes the ConvNet parameters. Our results in Table~\ref{tab:detection} indicate that \AlgoName{} performs comparably or better than state-of-the-art representation learning methods for these localization tasks.

\begin{table}[ht]
\caption{\textbf{Transfer learning: object detection and instance segmentation.} We benchmark learned representations on the object detection task on VOC07+12 using Faster R-CNN~\cite{ren2015faster} and on the detection and instance segmentation task on COCO using Mask R-CNN~\cite{he2017mask}. All methods use the C4 backbone variant~\cite{wu2019detectron2} and models on COCO are finetuned using the 1$\times$ schedule. Best results are in \textbf{bold}.}
\label{tab:detection}
\begin{center}
\setlength{\tabcolsep}{0.4em}\resizebox{\linewidth}{!}{
    \begin{tabular}{@{}lccccccccc@{}}
    \toprule
     Method & \multicolumn{3}{c}{VOC07+12 det} & \multicolumn{3}{c}{COCO det} & \multicolumn{3}{c}{COCO instance seg}\\
     \cmidrule(lr){2-4}\cmidrule(lr){5-7}\cmidrule(lr){8-10}
    & AP$_{\mathrm{all}}$ & AP$_{50}$ & AP$_{75}$ &  AP$^{\mathrm{bb}}$ & AP$^{\mathrm{bb}}_{50}$ & AP$^{\mathrm{bb}}_{75}$ & AP$^{\mathrm{mk}}$ & AP$^{\mathrm{mk}}_{50}$ & AP$^{\mathrm{mk}}_{75}$\\
    \midrule
    Sup. & 53.5 & 81.3 & 58.8 & 38.2 & 58.2 & 41.2 & 33.3 & 54.7 & 35.2 \\
    \midrule
    MoCo-v2 & \textbf{57.4} & 82.5 & \textbf{64.0} & \textbf{39.3} & 58.9 & \textbf{42.5} & \textbf{34.4} & 55.8 & 36.5 \\
    SwAV & 56.1 & \textbf{82.6} & 62.7 & 38.4 & 58.6 & 41.3 & 33.8 & 55.2 & 35.9 \\
    SimSiam & 57 & 82.4 & 63.7 & 39.2 & \textbf{59.3} & 42.1 & \textbf{34.4} & \textbf{56.0} & \textbf{36.7} \\
    BT (ours) & 56.8 & \textbf{82.6} & 63.4 & 39.2 & 59.0 & \textbf{42.5} & 34.3 & \textbf{56.0} & 36.5 \\
    \bottomrule
    \end{tabular}
}
\end{center}
\vskip -0.1in
\end{table}

\section{Ablations}

For all ablation studies, \AlgoName{} was trained for 300 epochs instead of 1000 epochs in the previous section. A linear evaluation on ImageNet of this baseline model yielded a $71.4\%$ top-1 accuracy and a $90.2\%$ top-5 accuracy. For all the ablations presented we report the top-1 and top-5 accuracy of training linear classifiers on the $2048$ dimensional \texttt{res5} features using the ImageNet train set.

\paragraph{Loss Function Ablations} 
We alter our loss function (eqn. \ref{eq:lossBarlow}) in several ways to test the necessity of each term of the loss function, and to experiment with different practices popular in other loss functions for SSL, such as \textsc{infoNCE}.  Table \ref{tab:abl-loss} recapitulates the different loss functions tested along with their results on a linear evaluation benchmark of Imagenet. First we find that removing the invariance term (on-diagonal term) or the redundancy reduction term (off-diagonal term) of our loss function leads to worse/collapsed solutions, as expected. We then study the effect of different normalization strategies. We first try to normalize the embeddings along the feature dimension so that they lie on the unit sphere, as it is common practice for losses measuring a cosine similarity \cite{chen2020simple,grill2020bootstrap,wang_understanding_2020}. Specifically, we first normalize the embeddings along the batch dimension (with mean subtraction), then normalize the embeddings along the feature dimension (without mean subtraction), and finally we measure the (unnormalized) covariance matrix instead of the (normalized) cross-correlation matrix in eqn. \ref{eq:crosscorr}. The performance is slightly reduced. Second, we try to remove batch-normalization operations in the two hidden layers of the projector network MLP. The performance is barely affected. Third, in addition to removing the batch-normalization in the hidden layers, we replace the cross-correlation matrix in eqn. \ref{eq:crosscorr} by the cross-covariance matrix (which means the features are no longer normalized along the batch dimension). The performance is substantially reduced. We finally try a cross-entropy loss with temperature, for which the on-diagonal term and off-diagonal term is controlled by a temperature hyperparameter $\tau$ and coefficient $\lambda$: 
$\mathcal{L} = -\log\sum_i\exp(\mathcal{C}_{ii}/\tau) + \lambda \log\sum_i\sum_{j\neq i}\exp(\max(\mathcal{C}_{ij}, 0)/\tau)$. The performance is reduced.



\begin{table}[ht]
\caption{\textbf{Loss function explorations}. We ablate the invariance and redundancy terms in our proposed loss and observe that both terms are necessary for good performance. We also experiment with different normalization schemes and a cross-entropy loss and observe reduced performance.}
\label{tab:abl-loss}
\vskip 0.15in
\begin{center}
\begin{tabular}{l c c}
\toprule
Loss function & Top-$1$ & Top-$5$ \\
\midrule
    Baseline & 71.4 & 90.2 \\
\midrule
    Only invariance term (on-diag term)  & 57.3 & 80.5 \\ 
    Only red. red. term (off-diag term)  & 0.1 & 0.5 \\
    \midrule
    Normalization along feature dim. & 69.8 & 88.8 \\
    No BN in MLP & 71.2 & 89.7 \\
    No BN in MLP + no Normalization & 53.4 & 76.7 \\
    \midrule
    Cross-entropy with temp. & 63.3 & 85.7 \\
\bottomrule
\end{tabular}
\end{center}
\vskip -0.1in
\end{table}

\begin{figure}[h!]
\vskip 0.2in
\begin{center}
\centerline{\includegraphics[width=8cm]{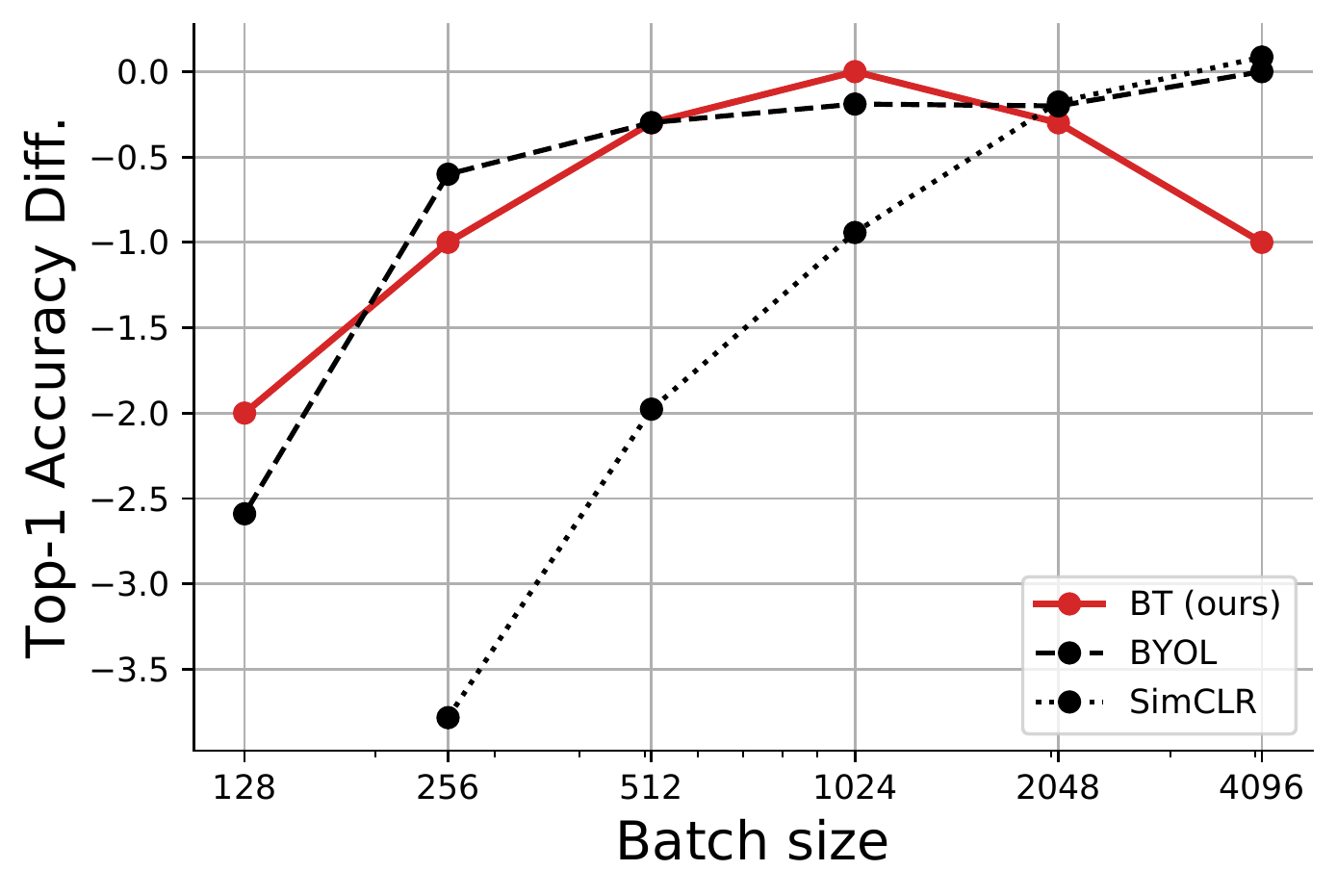}}
\caption{Effect of batch size. To compare the effect of the batch size across methods, for each method we report the difference between the top-1 accuracy at a given batch size and the best obtained accuracy among all batch size tested. \textsc{BYOL}: best accuracy is 72.5\% for a batch size of 4096 (data from \cite{grill2020bootstrap}  fig. 3A). \textsc{SimCLR}: best accuracy is 67.1\% for a batch size of 4096 (data from \cite{chen2020simple} fig. 9, model trained for 300 epochs). \AlgoName{}: best accuracy is 71.7\% for a batch size of 1024.}
\label{fig:fig_batch}
\end{center}
\vskip -0.2in
\end{figure}

\paragraph{Robustness to Batch Size} 
The \textsc{infoNCE} loss that draws negative examples from the minibatch suffer performance drops when the batch size is reduced (e.g. \textsc{SimCLR} \cite{chen2020simple}).  We thus sought to test the robustness of \AlgoName{} to small batch sizes. In order to adapt our model to different batch sizes, we performed a grid search on LARS learning rates for each batch size. We find that, unlike \textsc{SimCLR}, our model is robust to small batch sizes (Fig. \ref{fig:fig_batch}), with a performance almost unaffected for a batch as small as 256. In comparison the accuracy for SimCLR drops about $4$ p.p. for batch size 256. This robustness to small batch size, also found in non-contrastive methods such as \textsc{BYOL}, further demonstrates that our method is not only conceptually (see Discussion) but also empirically different than the \textsc{infoNCE} objective.

\begin{figure}[h!]
\vskip 0.2in
\begin{center}
\centerline{\includegraphics[width=8cm]{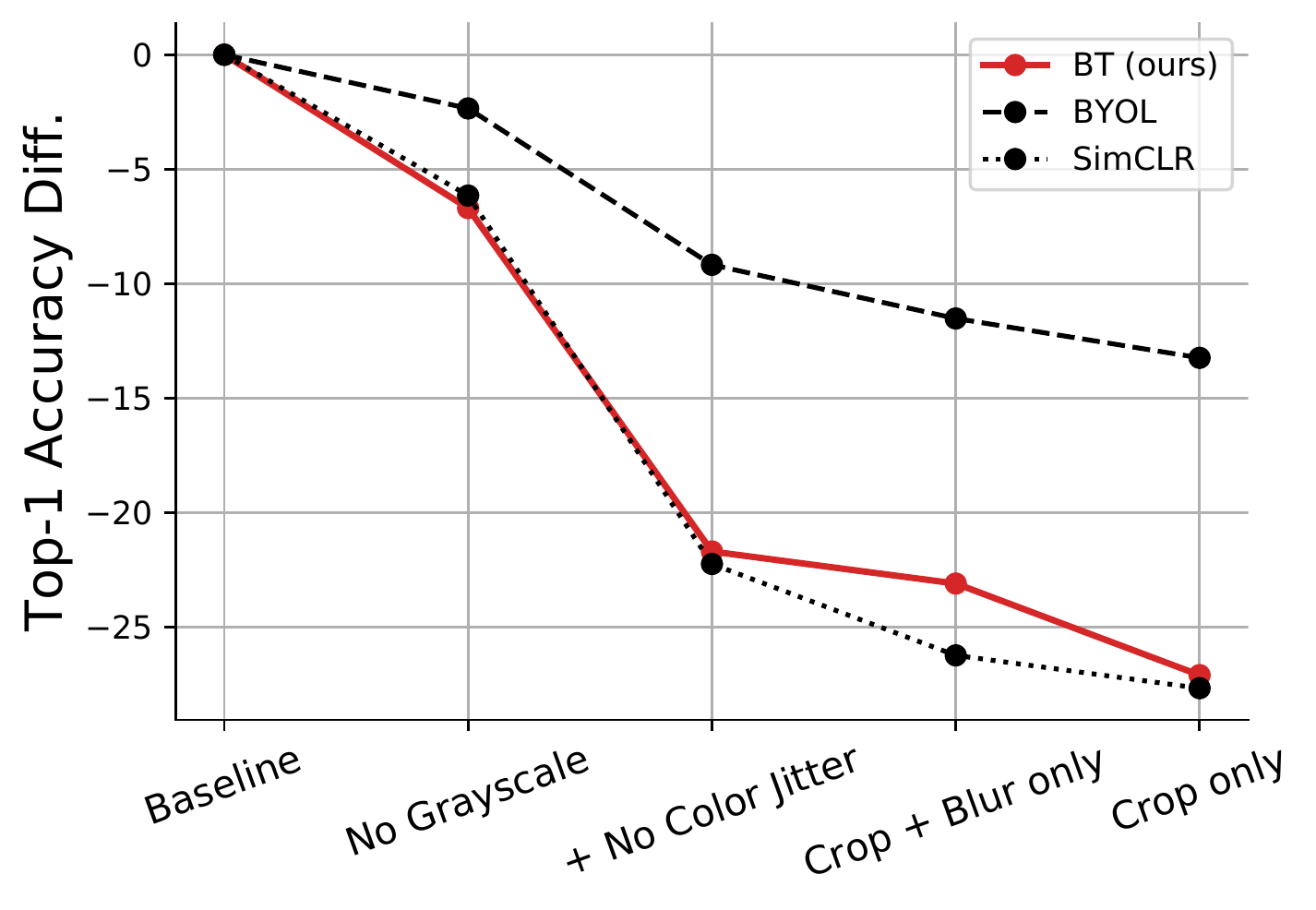}}
\caption{Effect of progressively removing data augmentations. Data for \textsc{BYOL} and \textsc{SimCLR} (repro) is from \cite{grill2020bootstrap} fig 3b. }
\label{fig:fig_augm}
\end{center}
\vskip -0.2in
\end{figure}

\paragraph{Effect of Removing Augmentations} 
We find that our model is not robust to removing some types of data augmentations, like \textsc{SimCLR} but unlike \textsc{BYOL} (Fig. \ref{fig:fig_augm}).  While this can be seen as a disadvantage of our method compared to \textsc{BYOL}, it can also be argued that the representations learned by our method are better controlled by the specific set of distortions used, as opposed to \textsc{BYOL} for which the invariances learned seem generic and intriguingly independent of the specific distortions used.

\begin{figure}[h!]
\vskip 0.2in
\begin{center}
\centerline{\includegraphics[width=8cm]{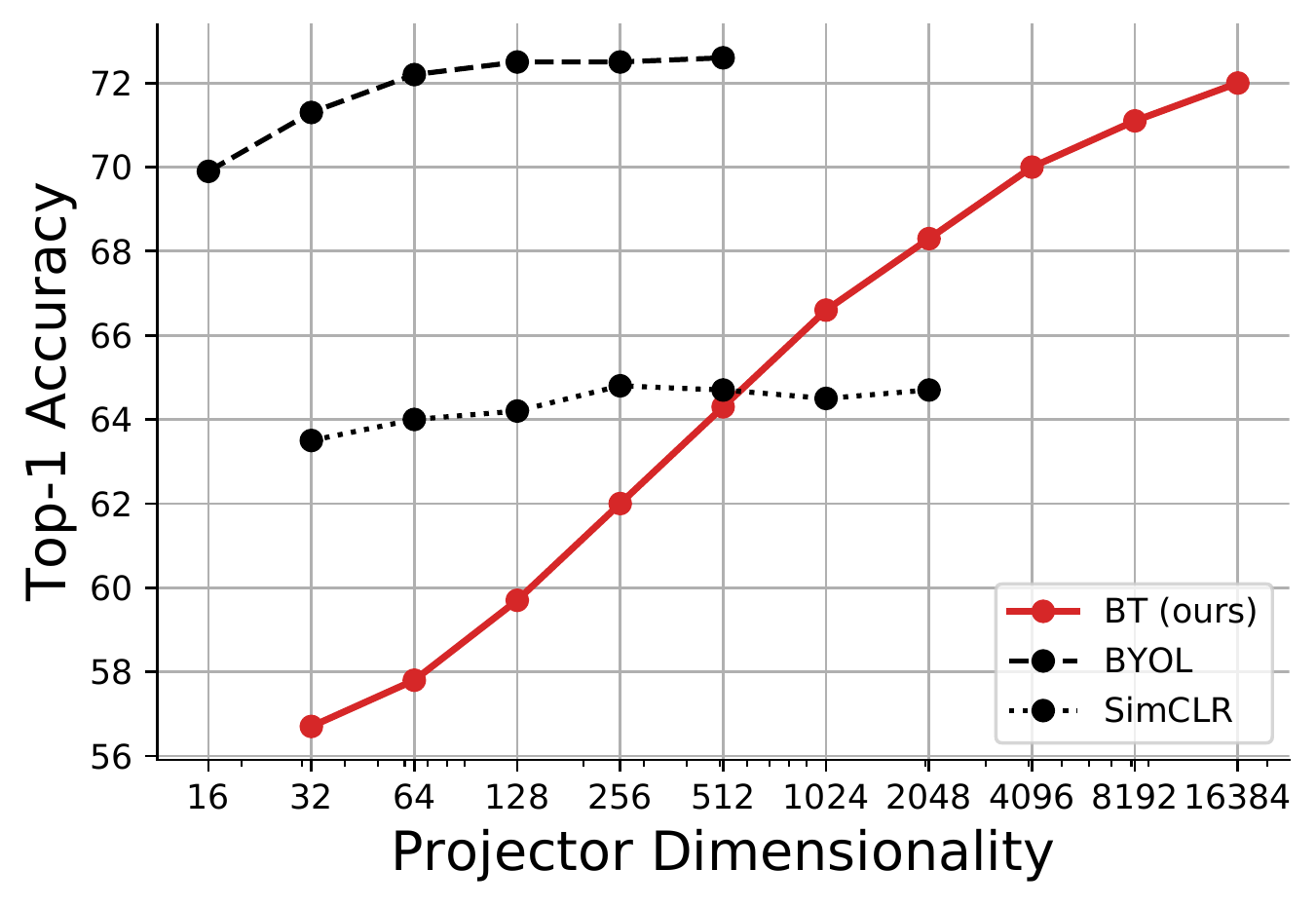}}
\caption{Effect of the dimensionality of the last layer of the projector network on performance. The parameter $\lambda$ is kept fix for all dimensionalities tested. Data for \textsc{SimCLR} is from \cite{chen2020simple} fig 8; Data for \textsc{BYOL} is from \cite{grill2020bootstrap} Table 14b.}
\label{fig:fig_projdim}
\end{center}
\vskip -0.2in
\end{figure}

\paragraph{Projector Network Depth \& Width} For other SSL methods, such as \textsc{BYOL} and \textsc{SimCLR}, the projector network drastically reduces the dimensionality of the ResNet output. In stark contrast, we find that \AlgoName{} performs better when the dimensionality of the projector network output is very large. Other methods rapidly saturate when the dimensionality of the output increases, but our method keeps improving with all output dimensionality tested (Fig. \ref{fig:fig_projdim}). This result is quite surprising because the output of the ResNet is kept fixed to 2048, which acts as a dimensionality bottleneck in our model and sets the limit of the intrinsic dimensionality of the representation. In addition, similarly to other methods, we find that our model performs better when the projector network has more layers, with a saturation of the performance for 3 layers.

\paragraph{Breaking Symmetry} Many SSL methods (e.g. \textsc{BYOL}, \textsc{SimSiam}, \textsc{SwAV}) rely on different symmetry-breaking mechanisms to avoid trivial solutions. Our loss function avoids these trivial solutions by construction, even in the case of symmetric networks. It is however interesting to ask whether breaking symmetry can further improve the performance of our network. Following \textsc{SimSiam} and \textsc{BYOL}, we experiment with adding a predictor network composed of 2 fully connected layers of size 8192 to one of the network (with batch normalization followed by a ReLU nonlinearity in the hidden layer) and/or a stop-gradient mechanism on the other network. We find that these asymmetries slightly decrease the performance of our network (see Table \ref{tab:abl-sym}). 

\begin{table}[h!]
\caption{Effect of asymmetric settings}
\label{tab:abl-sym}
\vskip 0.15in
\begin{center}
\begin{tabular}{ccccc}
\toprule
case & stop-gradient & predictor & Top-$1$ & Top-$5$ \\
\midrule
Baseline & - & - & 71.4 & 90.2 \\
\midrule
(a) & \checkmark & - & $70.5$ & $89.0$ \\
(b) & - & \checkmark & $70.2$ & $89.0$ \\
(c) & \checkmark & \checkmark & $61.3$ & $83.5$ \\
\bottomrule
\end{tabular}
\end{center}
\vskip -0.1in
\end{table}

\textbf{BYOL with a larger projector/predictor/embedding} For a fair comparison with BYOL, we also evaluated BYOL with a wider and/or deeper projector head (3-layer MLP), a wider and/or deeper predictor head, and a larger dimensionality of the embedding. BYOL did not improve under these conditions (see Table \ref{tab:BYOL_MLP3}).

\begin{table}[h!]
\caption{Wider and/or deeper projector and predictor heads and larger dimensionality of the embedding did not improve the performance of BYOL.\\}
\tiny
    \centering
    \begin{tabular}{l|l|c|l}
\hline
Projector&	Predictor&	Acc1	& Description \\
\hline
4096-256 &	4096-256 & 74.1\% &		baseline \\
4096-4096-256 &	4096-256 & 74.0\% &		3 layer proj,
2 layer pred,  256-d repr. \\
4096-4096-256	& 4096-4096-256	& 73.2\% &	3 layer proj, 3 layer pred, 256-d repr. \\
4096-4096-512 &	4096-512	& 73.7\% &	3 layer proj, 2 layer pred, 512-d repr. \\
4096-4096-512	& 4096-4096-512 & 73.2\% &		3 layer proj, 3 layer pred, 512-d repr. \\
8192-8192-8192	& 8192-8192 & 72.3\% &	same proj as BT, 2 layer pred, 8192-d repr. \\
\hline
    \end{tabular}
    \label{tab:BYOL_MLP3}
\end{table}

\begin{figure}[h!]
\vskip 0.2in
\begin{center}
\centerline{\includegraphics[width=8cm]{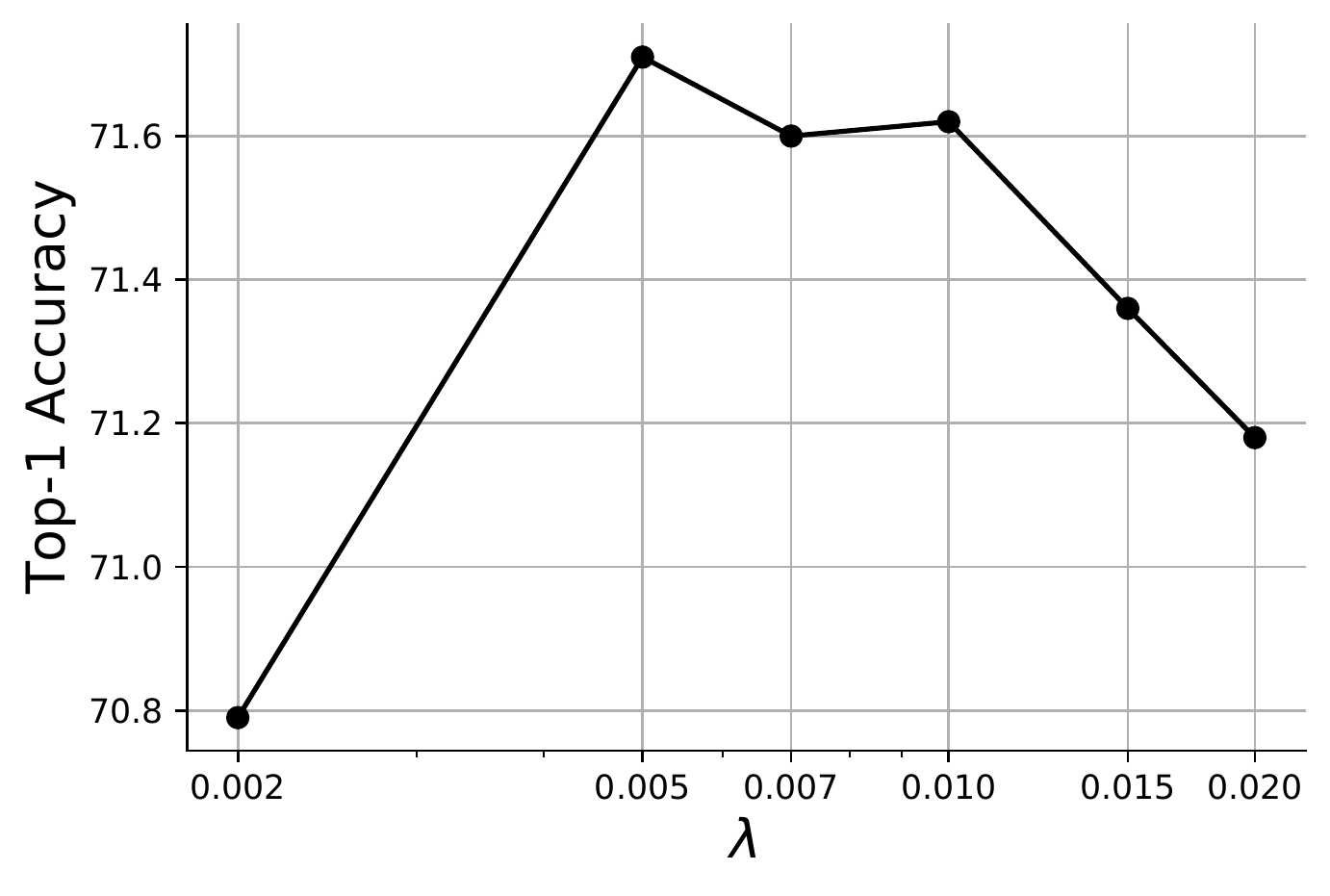}}
\caption{Sensitivity of \AlgoName{} to the hyperparameter $\lambda$}
\label{fig:fig_lambda}
\end{center}
\vskip -0.2in
\end{figure}

\textbf{Sensitivity to $\lambda$.} We also explored the sensitivity of \AlgoName{} to the hyperparameter $\lambda$, which trades off the desiderata of invariance and informativeness of the embeddings. We find that \AlgoName{} is not very sensitive to this hyperparameter (Fig. \ref{fig:fig_lambda}).

\section{Discussion}

\AlgoName{} learns self-supervised representations through a joint embedding of distorted images, with an objective function that maximizes similarity between the embedding vectors while reducing redundancy between their components. Our method does not require large batches of samples, nor does it require any particular asymmetry in the twin network structure. We discuss next the similarities and differences between our method and prior art, both from a conceptual and an empirical standpoint. For ease of comparison, all objective functions are recast with a common set of notations. The discussion ends with future directions.

\subsection{Comparison with Prior Art}

\paragraph{infoNCE}

The \textsc{InfoNCE} loss, where NCE stands for Noise-Contrastive Estimation \cite{gutmann_noise-contrastive_2010}, is a popular type of contrastive loss function used for self-supervised learning (e.g. \cite{oord2018representation, chen2020simple, he2019momentum,henaff2019data}). It can be instantiated as:
\begin{align*}
\mathcal{L}_{infoNCE} \triangleq - \underbrace{\sum_b \frac{\langle z^A_{b}, 
z^B_{b} \rangle_i}{\tau\norm{z^A_{b}}_2\norm{z^B_{b}}_2}}_\text{similarity term}\\
+ \underbrace{\sum_{b} \log\left( \sum_{b' \neq b} \exp \left(\frac{\langle z^A_{b}, 
z^B_{b'} \rangle_i}{\tau\norm{z^A_{b}}_2\norm{z^B_{b'}}_2}\right)\right)}_\text{contrastive term}
\end{align*}

where $z^A$ and $z^B$ are the twin network outputs, $b$ indexes the sample in a batch, $i$ indexes the vector component of the output, and $\tau$ is a positive constant called temperature in analogy to statistical physics.

For ready comparison, we rewrite \AlgoName{} loss function with the same notations:
\begin{align*}
\mathcal{L_{BT}} =  \underbrace{\sum_i \left(1-\frac{\langle z^A_{\boldsymbol{\cdot},i}, 
z^B_{\boldsymbol{\cdot},i} \rangle_b}{\norm{z^A_{\boldsymbol{\cdot},i}  }_2 \norm{z^B_{\boldsymbol{\cdot},i}}_2}\right) ^2}_\text{invariance term} \\
+ \lambda \underbrace{\sum_{i} \sum_{j \neq i} \left(\frac{\langle z^A_{\boldsymbol{\cdot},i}, 
z^B_{\boldsymbol{\cdot},j} \rangle_b}{\norm{z^A_{\boldsymbol{\cdot},i} }_2\norm{z^B_{\boldsymbol{\cdot},j}}_2}\right)^2}_\text{redundancy reduction term}
\end{align*}

Both \AlgoName{}' and \textsc{InfoNCE}'s objective functions have two terms, the first aiming at making the embeddings invariant to the distortions fed to the twin networks, the second aiming at maximizing the variability of the embedding learned.  Another common point between the two losses is that they both rely on batch statistics to measure this variability. However, the \textsc{InfoNCE} objective maximizes the variability of the embeddings by maximizing the pairwise distance between all pairs of samples, whereas our method does so by decorrelating the components of the embeddings vectors. 

The contrastive term in \textsc{InfoNCE} can be interpreted as a non-parametric estimation of the entropy of the distribution of embeddings \cite{wang_understanding_2020}. An issue that arises with non-parametric entropy estimators is that they are prone to the curse of dimensionality: they can only be estimated reliably in a low-dimensional setting, and they typically require a large number of samples. 

In contrast, our loss can be interpreted as a \emph{proxy} entropy estimator of the distribution of embeddings under \emph{a Gaussian parametrization} (see Appendix A). Thanks to this simplified parametrization, the variability of the embedding can be estimated from much fewer samples, and on very large-dimensional embeddings. Indeed, in the ablation studies that we perform, we find that (1) our method is robust to small batches unlike the popular \textsc{InfoNCE}-based method \textsc{SimCLR}, and (2) our method benefits from using very large dimensional embeddings, unlike \textsc{InfoNCE}-based methods which do not see a benefit in increasing the dimensionality of the output.

Our loss presents several other interesting differences with infoNCE:
\begin{itemize}
\item In \textsc{infoNCE}, the embeddings are typically normalized along the feature dimension to compute a cosine similarity between embedded samples. We normalize the embeddings along the batch dimension instead.

\item In our method, there is a parameter $\lambda$ that trades off how much emphasis is put on the invariance term vs. the redundancy reduction term. This parameter can be interpreted as the trade-off parameter in the \emph{Information Bottleneck} framework (see Appendix A). This parameter is not present in \textsc{infoNCE}. 

\item \textsc{infoNCE} also has a hyperparameter, the temperature, which can be interpreted as the width of the kernel in a non-parametric kernel density estimation of entropy, and practically weighs the relative importance of the hardest negative samples present in the batch \cite{chen2020simple}.
\end{itemize}

A number of alternative methods to ours have been proposed to alleviate the reliance on large batches of the \textsc{infoNCE} loss. For example, MoCo \cite{he2019momentum, chen2020improved} builds a dynamic dictionary of negative samples with a queue and a moving-averaged encoder. This enables building a large and consistent dictionary on-the-fly that facilitates contrastive unsupervised learning. MoCo typically needs to store $>60{,}000$ sample embeddings. In contrast, our method does not require such a large dictionary, since it works well with a relatively small batch size (e.g. 256).

\paragraph{Asymmetric Twins}

\textsc{Bootstrap-Your-Own-Latent} (aka \textsc{BYOL}) \cite{grill2020bootstrap} and \textsc{SimSiam} \cite{chen2020exploring} are two recent methods which use a simple cosine similarity between twin embeddings as an objective function, without \emph{any} contrastive term:

\begin{align*}
\mathcal{L}_{cosine} = - \sum_b \frac{\langle z^A_{b}, 
z^B_{b} \rangle_i}{\norm{z^A_{b}}_2\norm{z^B_{b}}_2}
\end{align*}

Surprisingly, these methods successfully avoid trivial solutions by introducing some asymmetry in the architecture and learning procedure of the twin networks. For example, \textsc{BYOL} uses a predictor network which breaks the symmetry between the two networks, and also enforces an exponential moving average on the target network weights to slow down the progression of the weights on the target network. Combined together, these two mechanisms surprisingly avoid trivial solutions. The reasons behind this success are the subject of recent theoretical and empirical studies \cite{tian_understanding_2020,chen2020exploring,fetterman_understanding_2020,richemond_byol_2020}. In particular, the ablation study \cite{chen2020exploring} shows that the moving average is not necessary, but that stop-gradient on one of the branch and the presence of the predictor network are two crucial elements to avoid collapse. Other works show that batch normalization \cite{tian_understanding_2020,fetterman_understanding_2020} or alternatively group normalization \cite{richemond_byol_2020} could play an important role in avoiding collapse.

Like our method, these asymmetric methods do not require large batches, since in their case there is no interaction between batch samples in the objective function.

It should be noted however that these asymmetric methods cannot be described as the optimization of an overall learning objective. Instead, there exists trivial solutions to the learning objective that these methods avoid via particular implementation choices and/or the result of non-trivial learning dynamics. In contrast, our method avoids trivial solutions by construction, making our method conceptually simpler and more principled than these alternatives (until their principle is discovered, see \cite{tian_understanding_2021} for an early attempt).

\paragraph{Whitening}
In a concurrent work, \cite{ermolov_whitening_2020} propose \textsc{W-MSE}. Acting on the embeddings from identical twin networks, this method performs a differentiable whitening operation (via Cholesky decomposition) of each batch of embeddings before computing a simple cosine similarity between the whitened embeddings of the twin networks. In contrast, the redundancy reduction term in our loss encourages the whitening of the batch embeddings as a soft constraint. The current W-MSE model achieves 66.3\% top-1 accuracy on the Imagenet linear evaluation benchmark. It is an interesting direction for future studies to determine whether improved versions of this hard-whitening strategy could also lead to state-of-the-art results on these large-scale computer vision benchmarks.

\paragraph{Clustering}
These methods, such as \textsc{DeepCluster} \cite{caron2018deep}, \textsc{SwAV} \cite{caron2020swav}, \textsc{SeLa} \cite{asano2019self}, perform contrastive-like comparisons without the requirement to compute all pairwise distances. Specifically, these methods simultaneously cluster the data while enforcing consistency between cluster assignments produced for different distortions of the same image, instead of comparing features directly as in contrastive learning. Clustering methods are also prone to collapse, \eg, empty clusters in k-means and avoiding them relies on careful implementation details. Online clustering methods like \textsc{SwAV} can be trained with large and small batches but require storing features when the number of clusters is much larger than the batch size. Clustering methods can also be combined with contrastive learning~\cite{li2020contrastive} to prevent collapse.

\paragraph{Noise As Targets} This method \cite{bojanowski2017unsupervised} learns to map samples to fixed random targets on the unit sphere, which can be interpreted as a form of whitening. This objective uses a single network, and hence does not leverage the distortions induced by twin networks. Predefining random targets might limit the flexibility of the representation that can be learned.


\paragraph{IMAX}

In the early days of SSL, \cite{becker_self-organizing_1992,zemel_discovering_1990} proposed a loss function between twin networks given by:
\begin{align*}
\mathcal{L}_{IMAX} \triangleq   \log |\mathcal{C}_{(Z^A-Z^B)}|-\log |\mathcal{C}_{(Z^A+Z^B)}|
\end{align*}
where $|~|$ denotes the determinant of a matrix, $\mathcal{C}_{(Z^A-Z^B)}$ is the covariance matrix of the difference of the outputs of the twin networks and  $\mathcal{C}_{(Z^A+Z^B)}$ the covariance of the sum of these outputs. It can be shown that this objective maximizes the information between the twin network representations under the assumptions that the two representations are noisy versions of the same underlying Gaussian signal, and that the noise is independant, additive and Gaussian. This objective is similar to ours in the sense that there is one term that encourages the two representations to be similar and another term that encourages the units to be decorrelated. However, unlike \textsc{IMAX}, our objective is not directly an information quantity, and we have an extra trade-off parameter $\lambda$ that trades off the two terms of our loss. The \textsc{IMAX} objective was used in early work so it is not clear whether it can scale to large computer vision tasks. Our attempts to make it work on ImageNet were not successful.

\subsection{Future Directions}

We observe a steady improvement of the performance of our method as we increase the dimensionality of the embeddings (i.e. of the last layer of the projector network). This intriguing result is in stark contrast with other popular methods for SSL, such as \textsc{SimCLR} \cite{chen2020simple} and \textsc{BYOL} \cite{grill2020bootstrap}, for which increasing the dimensionality of the embeddings rapidly saturates performance. It is a promising avenue to continue this exploration for even higher dimensional embeddings ($>16{,}000$), but this would require the development of new methods or alternative hardware to accommodate the memory requirements of operating on such large embeddings.

Our method is just one possible instanciation of the \emph{Information Bottleneck} principle applied to SSL. We believe that further refinements of the proposed loss function and algorithm could lead to more efficient solutions and even better performances. For example, the redundancy reduction term is currently computed from the off-diagonal terms of the cross-correlation matrix between the twin network embeddings, but alternatively it could be computed from the off-diagonal terms of the auto-correlation matrix of a single network's embedding. Our preliminary analyses seem to indicate that this alternative leads to similar performances (not shown). A modified loss could also be applied to the (unnormalized) cross-covariance matrix instead of the (normalized) cross-correlation matrix (see Ablations for preliminary analyses).

\section*{Acknowledgements}
We thank Pascal Vincent, Yubei Chen and Samuel Ocko for helpful insights on the mathematical connection to the infoNCE loss, Robert Geirhos and Adrien Bardes for extra analyses not included in the manuscript and Xinlei Chen, Mathilde Caron, Armand Joulin, Reuben Feinman and Ulisse Ferrari for useful comments on the manuscript.

\bibliography{refs}
\bibliographystyle{icml2021}
\newpage
\appendix


\section{Connection between \AlgoName{} and the Information Bottleneck Principle}

\begin{figure}[ht]
\vskip 0.2in
\begin{center}
 \centerline{\includegraphics[width=6cm]{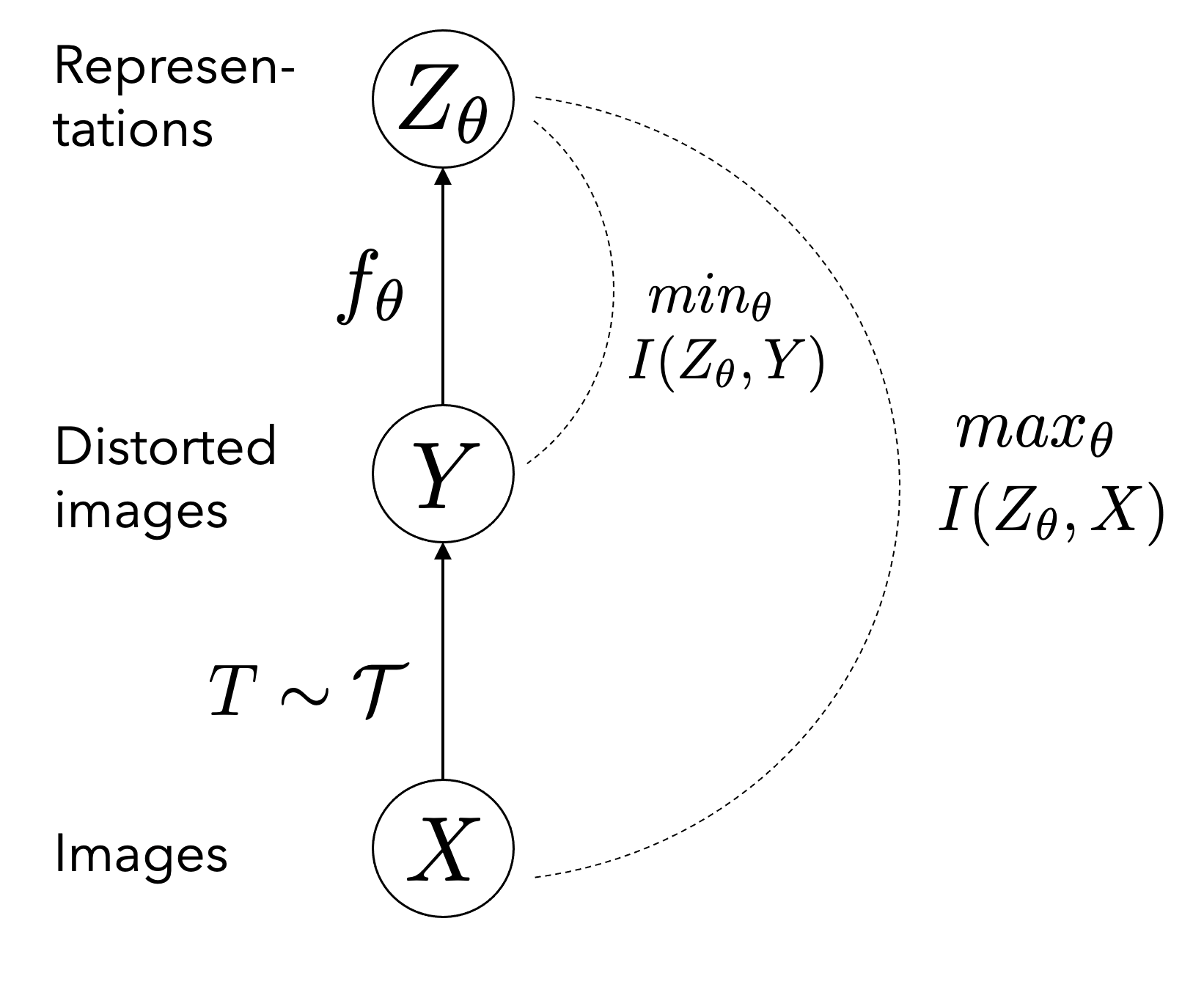}}
\caption{The information bottleneck principle applied to self-supervised learning (SSL) posits that the objective of SSL is to learn a representation $Z_\theta$ which is informative about the image sample, but invariant (i.e. uninformative) to the specific distortions that are applied to this sample. \AlgoName{} can be viewed as a specific instanciation of the information bottleneck objective.}
\label{fig:fig_IB}
\end{center}
\vskip -0.2in
\end{figure}

We explore in this appendix the connection between \AlgoName{}' loss function and the \emph{Information Bottleneck} (IB) principle \cite{tishby_deep_2015,tishby_information_2000}. 

As a reminder, \AlgoName{}' loss function is given by:

\begin{equation}
\mathcal{L_{BT}} \triangleq  \underbrace{\sum_i  (1-\mathcal{C}_{ii})^2}_\text{invariance term}  + ~~\lambda \underbrace{\sum_{i}\sum_{j \neq i} {\mathcal{C}_{ij}}^2}_\text{redundancy reduction term}
\label{eq:lossBarlow_app}
\end{equation}

where $\lambda$ is a positive constant trading off the importance of the first and second terms of the loss, and where $\mathcal{C}$ is the cross-correlation matrix computed between the outputs of the two identical networks along the batch dimension :

\begin{equation}
\mathcal{C}_{ij} \triangleq \frac{
\sum_b z^A_{b,i} z^B_{b,j}}
{\sqrt{\sum_b {(z^A_{b,i})}^2} \sqrt{\sum_b {(z^B_{b,j})}^2}}
\label{eq:crosscorr_app}
\end{equation}

where $b$ indexes batch samples and $i,j$ index the vector dimension of the networks' outputs. $\mathcal{C}$ is a square matrix with size the dimensionality of the network's output, and with values comprised between -1 (i.e. perfect anti-correlation) and 1 (i.e. perfect correlation). 

Applied to self-supervised learning, the IB principle posits that a desirable representation should be as informative as possible about the sample represented while being as invariant (i.e. non-informative) as possible to distortions of that sample (here the data augmentations used) (Fig. \ref{fig:fig_IB}). This trade-off is captured by the following loss function:

\begin{equation}
   \mathcal{IB_{ \theta}} \triangleq I(Z_{\theta}, Y) - \beta I(Z_{\theta}, X)
    \label{eq:lossIB}
\end{equation}

where $I(.,.)$ denotes mutual information and $\beta$ is a positive scalar trading off the desideratas of preserving information and being invariant to distortions.

Using a classical identity for mutual information, we can rewrite equation \ref{eq:lossIB} as:

\begin{equation}
   \mathcal{IB_{ \theta}} = [H(Z_{\theta}) - \cancelto{0}{H(Z_{\theta}|Y)}] - \beta [H(Z_{\theta}) - H(Z_{\theta}|X)] 
    \label{eq:lossIB2}
\end{equation}

where $H(.)$ denotes entropy. The conditional entropy $H(Z_{\theta}|Y)$ ---the entropy of the representation conditioned on a specific distorted sample--- cancels to 0 because the function $f_{\theta}$ is deterministic, and so the representation $Z_{\theta}$ conditioned on the input sample $Y$ is perfectly known and has zero entropy. Since the overall scaling factor of the loss function is not important, we can rearrange equation \ref{eq:lossIB2} as: 

\begin{equation}
   \mathcal{IB_{ \theta}} =  H(Z_{\theta}|X) + \frac{1-\beta}{\beta}  H(Z_{\theta})
    \label{eq:lossIB3}
\end{equation}

Measuring the entropy of a high-dimensional signal generally requires vast amounts of data, much larger than the size of a single batch. In order to circumvent this difficulty, we make the simplifying assumption that the representation $Z$ is distributed as a Gaussian. The entropy of a Gaussian distribution is simply given by the logarithm of the determinant of its covariance function (up to a constant corresponding to the assumed discretization level that we ignore here) \cite{cai_law_2015}. The loss function becomes:  

\begin{equation}
   \mathcal{IB_{ \theta}} =  ~\mathbb{E}_{X} log ~|\mathcal{C}_{Z_\theta|X}| + \frac{1-\beta}{\beta} ~log ~|\mathcal{C}_{Z_\theta}|
    \label{eq:lossIB4}
\end{equation}

This equation is still not exactly the one we optimize for in practice (see eqn. \ref{eq:lossBarlow_app} and \ref{eq:crosscorr_app}). Indeed, our loss function is only connected to the IB loss given by eqn. \ref{eq:lossIB4} through the following simplifications and approximations:
\begin{itemize}
  \item In the case where $\beta<=1$, it is easy to see from eqn. \ref{eq:lossIB4} that the best solution to the IB trade-off is to set the representation to a constant that does not depend on the input. This trade-off thus does not lead to interesting representations and can be ignored. When $\beta>1$, we note that the second term of eqn. \ref{eq:lossIB4} is preceded by a negative constant. We can thus simply replace $\frac{1-\beta}{\beta}$ by a new positive constant $\lambda$, preceded by a negative sign.
  
  \item In practice, we find that directly optimizing the determinant of the covariance matrices does not lead to SoTA solutions. Instead, we replace the second term of the loss in eqn. \ref{eq:lossIB4} (maximizing the information about samples), by the proxy which consist in simply minimizing the Frobenius norm of the cross-correlation matrix. If the representations are assumed to be re-scaled to 1 along the batch dimension before entering the loss (an assumption we are free to make since the cross-correlation matrix is invariant to this re-scaling), this minimization only affects the off-diagonal terms of the covariance matrix
  (the diagonal terms being fixed to 1 by the re-scaling) and encourages them to be as close to 0 as possible. It is clear that this surrogate objective, which consists in decorrelating all output units, has the same global optimum than the original information maximization objective. 
  
  \item For consistency with eqn. \ref{eq:lossIB4}, the second term in \AlgoName' loss should be computed from the auto-correlation matrix of one of the twin networks, instead of the cross-correlation matrix between twin networks. In practice, we do not see a strong difference in performance between these alternatives.
  
  \item Similarly, it can easily be shown that the first term of eqn. \ref{eq:lossIB4} (minimizing the information the representation contains about the distortions) has the same global optimum than the first term of eqn. \ref{eq:lossBarlow_app}, which maximizes the alignment between representations of pairs of distorted samples.  
\end{itemize}

\section{Evaluations on ImageNet}

\subsection{Linear evaluation on ImageNet}
\label{sec:linear_evaluation}
The linear classifier is trained for 100 epochs with a learning rate of $0.3$ and a cosine learning rate schedule. We minimize the cross-entropy loss with the SGD optimizer with momentum and weight decay of $10^{-6}$. We use a batch size of $256$. At training time we augment an input image by taking a random crop, resizing it to $224 \times 224$, and optionally flipping the image horizontally. At test time we resize the image to $256 \times 256$ and center-crop it to a size of $224 \times 224$.

\subsection{Semi-supervised training on ImageNet}
\label{sec:semisupervised_evaluation}
We train for $20$ epochs with a learning rate of $0.002$ for the ResNet-50 and $0.5$ for the final classification layer. The learning rate is multiplied by a factor of $0.2$ after the 12th and 16th epoch. We minimize the cross-entropy loss with the SGD optimizer with momentum and do not use weight decay. We use a batch size of $256$. The image augmentations are the same as in the linear evaluation setting.

\section{Transfer Learning}

\subsection{Linear evaluation} We follow the exact settings from PIRL~\cite{misra2019self} for evaluating linear classifiers on the Places-205, VOC07 and iNaturalist2018 datasets. For Places-205 and iNaturalist2018 we train a linear classifier with SGD (14 epochs on Places-205, 84 epochs on iNaturalist2018) with a learning rate of $0.01$ reduced by a factor of $10$ at two equally spaced intervals, a weight decay of $5\times10^{-4}$ and SGD momentum of $0.9$. We train SVM classifiers on the VOC07 dataset where the $C$ values are computed using cross-validation.

\subsection{Object Detection and Instance Segmentation}
We use the detectron2 library~\cite{wu2019detectron2} for training the detection models and closely follow the evaluation settings from~\cite{he2019momentum}. The backbone ResNet50 network for Faster R-CNN~\cite{ren2015faster} and Mask R-CNN~\cite{he2017mask} is initialized using our \AlgoName{} pretrained model.

\par \noindent \textbf{VOC07+12} We use the VOC07+12 \texttt{trainval} set of $16K$ images for training a Faster R-CNN~\cite{ren2015faster} C-4 backbone for $24K$ iterations using a batch size of $16$ across $8$ GPUs using SyncBatchNorm. The initial learning rate for the model is $0.1$ which is reduced by a factor of $10$ after $18K$ and $22K$ iterations. We use linear warmup~\cite{goyal2017accurate} with a slope of $0.333$ for $1000$ iterations.
\par \noindent \textbf{COCO} We train Mask R-CNN~\cite{he2017mask} C-4 backbone on the COCO 2017 \texttt{train} split and report results on the \texttt{val} split. We use a learning rate of $0.03$ and keep the other parameters the same as in the $1\times$ schedule in detectron2.


\end{document}